# Knowledge-Based Decision Model Construction for Hierarchical Diagnosis: A Preliminary Report

**Soe-Tsyr Yuan**
Computer Science Department
Oregon State University
Corvallis, OR 97331

## Abstract

Numerous methods for probabilistic reasoning in large, complex belief or decision networks are currently being developed. There has been little research on automating the dynamic, incremental construction of decision models. A uniform value-driven method of decision model construction is proposed for the hierarchical complete diagnosis. Hierarchical complete diagnostic reasoning is formulated as a stochastic process and modeled using influence diagrams. Given observations, this method creates decision models in order to obtain the best actions sequentially for locating and repairing a fault at minimum cost. This method construct decision models incrementally, interleaving probe actions with model construction and evaluation. The method treats meta-level and base-level tasks uniformly. That is, the method takes a decision-theoretic look at the control of search in causal pathways and structural hierarchies.

## 1   Introduction

Over the past several years, there has been great interest in AI concerning decision-theoretic methods for uncertain reasoning and decision making [2, 10, 9]. Using these methods, researchers have brought the advantages of mathematical clarity and a well-founded normative basis to AI problems involving choice and uncertainty. However, these methods require specific decision model languages, for example, influence diagrams, to express and reason with the available knowledge. These model languages are inadequate for expressing general relationships among concepts and therefore unsuitable for basic knowledge representation in large scale problem domains. They are, however, ideally suitable for reasoning about particular problem situations. Therefore, a domain knowledge base must be encoded in a general-purpose source language. Decision models in target language can then be dynamically generated for a particular problem instance encountered. We refer this approach as knowledge-based decision model construction (KB-DMC) [3].

Current KBDMC systems are problem-characteristic dependent. In most previous KBDMC systems [2, 18, 7], the incremental construction of models is an off-line process. That is, these systems do not perform active information gathering in generating networks from knowledge bases. This paper provides an on-line KBDMC system and first uses the technique of top-down hierarchical incremental construction of decision models.

Our KBDMC method is applied to the problems of *resource-limited, hierarchical, complete diagnostic reasoning*. These problems are resource-limited because their diagnosis plans are severely constrained by minimum cost, hierarchical because their diagnosis domains encode *functional subsystem part-of hierarchies*, and complete because their diagnosis plans take into account the complete path - from observation, to hypothesized diagnosis, to treatments for the diagnosis [13]. For example, when diagnosing any circuit or mechanical system in which there is a functional subsystem part-of hierarchy, our goal is to locate and repair the defect in the device at minimum cost. Furthermore, we use the concept of *causal pathways* [5] as a primary component of the knowledge needed to do diagnostic reasoning from structure and behavior. These causal pathways specify how one component affects another, indicating *categories of failure*. For example, the functional pathway models functional errors of components and the bridge fault pathway models a class of wiring errors between components. Careful organization of causal pathways allows us to make simplifying initial assumptions, surrendering them gracefully to consider more complex hypotheses when necessary. In this paper, we use a combinatorial circuit as our problem domain example and consider two paths of causal interaction (functional and bridge fault).

When an autonomous agent operates in a resource-limited environment, this agent's plans will typically be severely constrained by the limitations of time, cost, or other critical resources. Furthermore, an agent's knowledge of the world is always incomplete and subject to change. The agent must be able to deal with every kind of uncertainty (e.g., information uncertainty and control uncertainty) in its knowledge of the world. Therefore, we formulate our diagnostic model as a stochastic process. In this stochastic process, the information uncertainty includes prior probabilities of failure of device components or chips, link probabilities, and analytical probability information regarding the failure of causal pathways; the control uncertainty includes the optimal selection of causal pathway and optimal selection of actions in a causal pathway. Here, optimum is with respect to minimum cost, which includes the external repair cost and internal computational cost. Furthermore, this stochastic process is modeled using influence diagrams [8]. Influence diagrams are graphical knowledge representations for decision problem instances under uncertainty. Influence diagrams are well-defined, formalized decision networks for which evaluation algorithms [14, 4, 17, 15] have been developed. Evaluation of influence diagrams gives us the optimal policy of the stochastic problem instance with respect to the decision maker preferences. That is, our method decision-theoretically formulates the selection of a causal pathway to model, focus of attention within a pathway, and base level actions.

Our method presumes a problem description including domain knowledge expressed in a source language and a problem instance described as a set of observations. Our KBDMC procedure maps observations into a case-specific decision model expressed in our target language (influence diagrams). The method then evaluates the resulting decision model to identify the decisions to locate and repair the fault for the given diagnosis problem instance.

This paper is organized as follows. First, we briefly describe the skeleton of the proposed method. Then, we present and discuss the components of the method. We demonstrate how the method is applied to a circuit instance example. We compare our work with the other related work in diagnosis. We conclude with a discussion of the advantages of the method, and present the future work.

## 2  Skeleton of our Methodology

In this section we briefly describe our source and target languages. We then describe the general principles underlying our KBDMC method.

We use as our source language an object-centered representation which is based on the terminological part of *KL-ONE* [1]. We use as our target language influence diagrams (ID), which are networks with directed arcs and no cycles. The nodes of an ID represent random variables, decisions, and preferences (captured in a distinguished *Value* node). Arcs into random variables indicate probabilistic dependence, and arcs into decision nodes specify the information available at the time of decision-making. The evaluation process of an ID is driven by the goal of maximizing expected utility.

The general diagnostic principle we use is uniform decision-theoretic modeling of both meta-level control decisions among causal pathways and base-level control decisions within a causal pathway. In this paper we use a circuit example and consider two types (causal pathways) of errors (device component and bridge fault errors). Consequently, there are three components to the application of our method to this problem: a meta-level component, a functional component, and a bridge fault component. The meta-level component formulates the decision model to determine which causal pathway to explore. The functional component constructs the decision model to obtain fault hypotheses and corresponding actions within the functional causal pathway. The bridge fault component formulates the decision model to obtain the actions and the fault hypotheses within the bridge fault causal pathway. Our method begins with the meta-level component in order to choose the causal pathway to explore first. Our method repeats this process whenever a selected causal pathway component reports failure. The causal pathway components declare failure whenever they cannot locate the fault in the corresponding layer.

In the functional component, our method searches for variables to be included in a decision model incrementally. This search is driven by the hierarchical functional structure of the device, the functional causal pathways within the device, the initial observations, and the data gathered through probe actions as the decision model is elaborated.

Our method is based on the following assumptions: (1) single fault; (2) complete domain knowledge; (3) all the probabilistic information provided; (4) static (although not necessarily uniform) probe costs.

## 3  The Functional Component

The main idea behind the functional component is to progressively elaborate a conceptual decision model (shown in figure 1). This elaboration is a top-down hierarchical refinement without backtracking, and is guided by the functional characteristics of the device (ie, a well defined functional subsystem hierarchy). Note that throughout this paper we assume a subsystem is ok if and only if it produces the correct output for any given random input (that is, for all inputs).

We use an *one step computational conceptual decision model* as shown in figure 1, which is stored in the conceptual-id object. There are two important types of actions which are central to this approach:

- **Goal-achievement actions:** actions which directly satisfy the agent's goal, such as the Treat-

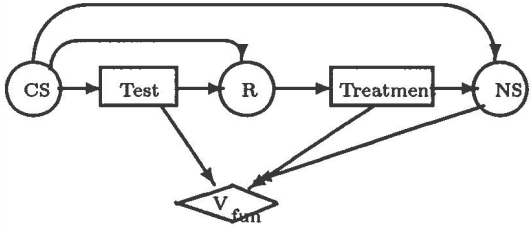

Figure 1: Conceptual decision model.

ment decision node in figure 1.

- **Information-gathering actions:** actions which reduce the agent's uncertainty by gathering new information, such as the Test decision node in figure 1.

In each computation step, we consider only one information gathering action (Test) and one goal-achievement action (Treatment) .

We will use the following terms in describing the functional component:

- **Context:** A focus of attention in the device. It includes a set of subsystems or components, explicitly represented in a functional-component decision model.
- **CS:** Current state of current Context.
- **NS:** Next state of current Context.
- **Test:** Decision regarding which testpoint should be probed.
- **R:** Result of Test.
- **Treatment:** Decision regarding which treatment should be taken.
- $V_{fun}$ : The **Value function** which guides model elaboration within the functional component is:

$$C(test)+C(treatment)+C(statusfollowingtreatment)$$

where C indicates the cost function.

### 3.1    Functional Component Procedure

The incremental top-down hierarchical elaboration of decision model is accomplished by the following steps:

1. **Refine the conceptual decision model:** In the conceptual decision model, the decision nodes Test and Treatment in figure 1 are completely dependent on CS and NS. When working with large knowledge bases, explicitly representing the entire device in full detail results in an intractable ID. Therefore, we will build models dynamically and incrementally, scoping the current Context to maintain tractability. Our method for scoping the Context is a top-down technique performed with the aid of the functional subsystem hierarchy.

(a) Substitute CS and NS with the highest level faulted subsystem hierarchy.

(b) Repeat *functional meta process* until this process arrives at a *functional base process* or stops at a middle Functional meta process.

### Functional meta process:

**Definition :**    Define a functional meta process to be a decision process like figure 1, in which the most recent leaves of the faulted subsystem hierarchy tree in CS and NS include at least one non-component subsystem object. (Hereafter, the most recent expanded leaves of the faulted subsystem hierarchy tree of CS and NS are referred to as the *current context of current functional meta process*.) Each functional meta process updates the content of the conceptual-id object, which store the most recent decision model constructed.

### Functional meta process procedure:

1. **Define decision alternatives:**

   - **testpoints:** The set of current relevant testpoints is the set of output points of the subsystems in the current context of current functional meta process. The decision node Test has as alternatives measuring the value of each of these testpoints.
   - **treatments for base level components:** Each base level component has a set of predefined treatments and their corresponding costs. For example, the treatments for a base level component NOR include "nothing" and "replace".
   - **treatments for subsystem components:** Each subsystem has treatments such as "nothing", "replace", and "repair". "Nothing" and "replace", like the treatments of the base level components, are predefined and have their corresponding costs stored in a knowledge base. However, the cost of a "repair" treatment is evaluated dynamically and will be described later.

     The decision node Treatment has as alternatives the treatment options on the base level components and subsystems in the current context of current functional meta process.

2. **Compute the costs of repair alternatives:** The repair treatment for a subsystem can be thought of as a means for focusing expansion. That is, when the result of the decision node Treatment is made to be "repair" on a subsystem, then our method will expand this subsystem and check one step further (that is, to add the children of this subsystem into CS and NS), narrowing down the Context.

   Therefore, before we make a treatment decision, we must compute the costs of the repair treatments of all subsystems in the current context of current functional meta process. We have a *complete technique* to get the exact cost value and an *incomplete heuristic technique* to estimate the cost value of the repair treatment of a subsystem. The complete technique uses the way of bottom-up and level-order representing the knowledge of the functional subsystem hierarchy in the source language, lets the repair cost for each leave-level component node be its replacement cost, and lets the repair cost of each intermediate subsystem node be the summation of its inspection cost and the minimum repair cost among its children. The incomplete technique expands the hierarchy till a fixed depth horizon from the subsystem, in each branch, calculates the summation of the accumulated inspection cost till the horizon and the replacement cost of the horizon node, backups the values computed to the subsystem with the minimum criteria, and chooses the minimum value as the estimate cost value of the repair treatment for the subsystem.

3. **Evaluate the decision model.**

4. **Execute the recommended actions:**

   - - Repair - Expand the selected subsystem and let the context of next functional meta process be the children of this selected subsystem. That is, subsequent evaluation of the decision model will only consider the newly added expansions: the links coming out of the Treatment decision node are only connected to the newly added nodes in NS, the links from CS to R only consider the newly added portion in CS, and the links between CS and NS only consider the newly added portion of expansion. The reason for this updating is that we only have to consider the newly added nodes, it is unnecessary to re-evaluate the nodes we have already evaluated (under assumptions of fault uniqueness and stability).

   - - Replace - If this process makes a nothing or replacement decision on a subsystem, then after executing recommended actions, the functional component observes device I/O to determine system status. If the device still fails to work, then this process declares failure.

**Functional base process:**

**Definition:** Define a functional base process to be a decision process like figure 1 in which the most recent expanded leaves of the faulted subsystem hierarchy tree of CS and NS include only base level component objects.

**Functional base process procedure:**

1. **Evaluate the decision model.**

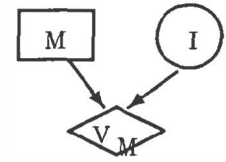

Figure 2: Meta-level decision model.

2. **Execute the recommended action:** Just execute the recommended nothing or replacement action on a component. If the device fails to work, then this process declares failure.

# 4   The Bridge fault component

We model bridge faults as occurring across adjacent pins at a chip. The bridge fault component uses probability information concerning potential bridge faults at chips to formulate a decision model over which chip to test next and then tests this chip, using physical packaging information in combination with functional pathway information. (We omit detail description due to page limitations.)

# 5   The Meta-Level Component

The meta-level component uses information on the probabilities of the failure in causal pathways and estimated time complexities of functional and bridge fault components to formulate a decision model over which causal pathway to explore next. Our procedure assumes the following:

1. **A template decision model as in figure 2.**

2. **M:** choice of a causal interaction which can be FL (functional component) or BFL (bridge fault component).

3. **I:** uncertain variable describing the actual fault causal pathway.

4. $V_M$: Value function as follows :

OUTCOME SPACE

| M | I | $V_M$ |
|------|------|-------|
| FL | FL | $X_1*u+X_2$ |
| FL | BFL | $X_1*u+X_2+Y_1*u+Y_2$ |
| BFL | FL | $Y_1*u+Y_2+X_1*u+X_2$ |
| BFL | BFL | $Y_1*u+Y_2$ |

- $X_1$: expected time complexity for the functional component lookahead.

- $X_2$: replacement cost and expected total inspection cost in the functional component lookahead. (We assume there is an inspection cost when our method decides to expand a subsystem one step further.)

- $Y_1$: expected time complexity for the bridge fault component lookahead.

- $Y_2$: final technician effort cost in the bridge fault component lookahead.
- u: cost of a faulted device per unit time.

### 5.1  Meta Level Decision Procedure

Before we evaluate the template decision model to obtain the actions, we perform the following steps to complete the value function of this decision model.

1. **Identify functional component lookahead:**

   **Definition:** Define the *current horizon of the functional component* to be the leaves of the subtree rooted at the same parent node as node A, where node A is a node at which the functional component has reported failure, or the initial highest level faulted subsystem node if the functional component has not yet been executed.

   **Functional component lookahead:**
   If the functional component has previously been executed and reported failure, the current Context will be modified, and then the functional component will be restarted. The process of the functional component lookahead will be executed if the meta-level component choose it.

   (a) If the functional component has not yet executed, initialize with a decision model containing the highest level faulted device subsystems and proceeds to step (d).

   (b) Prune node A and its associated links in the current functional component decision model.

   (c) Within the scope of the faulted subsystem hierarchy, let the sibling of node A with the same parent (node B) be the current horizon of the functional component lookahead. If node A is the only child of its parent, then prune the whole subtree rooted at node B, and let the sibling of node B with the same parent (node C) be the current horizon of the functional component lookahead.

   (d) Step (b) in functional component procedure as in 3.1.

2. **Compute the expected time complexity and external repair cost for the functional component:**

   **Assumptions:**  In order to make the analysis more tractable, we make the following assumptions :

   - $b_{avg}$: average number of the branches in the subsystem hierarchy.
   - d: maximum depth of the functional subsystem hierarchy.
   - $d_{max}$: maximum depth from the current new horizon to the base level in the subsystem hierarchy.

- $f_{eval}$: procedure which outputs the estimated number of multiplication operations required for evaluating a given influence diagram. (This procedure first translates an influence diagram into a belief net (BN) and then uses the SPI partition method [11] to estimate the number of multiplications required .)
- $ID_i$: expected influence diagram of the current $i$th functional meta process. It is derived from the current $ID_{i-1}$ by adding $b_{avg}$ nodes and associated links to a node of the horizon of the current $ID_{i-1}$. $ID_0$ is the initial content of the conceptual-id object. (Here we have omitted some detailed assumptions for simplicity.)
- $id_i$  $f_{eval}(ID_i)$, estimated number of multiplication operations required for evaluating the current expected influence diagram $ID_i$.

$X_1$  Expected time complexity for the functional component lookahead:
The functional component will build and evaluate a sequence of progressively refined decision models through a successive of functional meta processes and a functional base process. The expected time complexity of this process is estimated as follows:

- In each functional meta process, we assume the time complexity to compute the repair treatment costs and the time complexity to expand and modify an influence diagram are constant, once $b_{avg}$ is assigned and stays fixed.
- Total expected time complexity of the functional component lookahead is calculated by the following :
  After the the process of functional component lookahead arrives at a new horizon, it may require between one and $d_{max}+1$ meta processes to obtain the optimal solution. We assume that there is uniform distribution among the following cases:

  i meta process  $\sum_{k=0}^{i-1} id_{d-d_{max}+k}$, i=1 to $d_{max}+1$.
  Hence, the final expected time complexity is
  $id_{d_{max}} + d_{max}/(d_{max}+1)*id_{d_{max}+1} + (d_{max}-1)/(d_{max}+1)*id_{d-d_{max}+2}$                +
  $..+1/(d_{max}+1)*id_d$.

$X_2$  External repair cost (replacement cost and expected total inspection cost) :

- Expand the current horizon until a base level is reached.
- At each node, calculate the current accumulated inspection cost from the horizon and then add it to the current replacement cost. We call this summation value $x_2$.
- At each level, calculate the average corresponding $x_2$ value, say $x_2^i$, where i represents the execution of i functional meta processes from the current horizon, and after these i functional meta processes, our method will stop at the $(d - d_{max} + i - 1)$th level of the functional subsystem hierarchy.

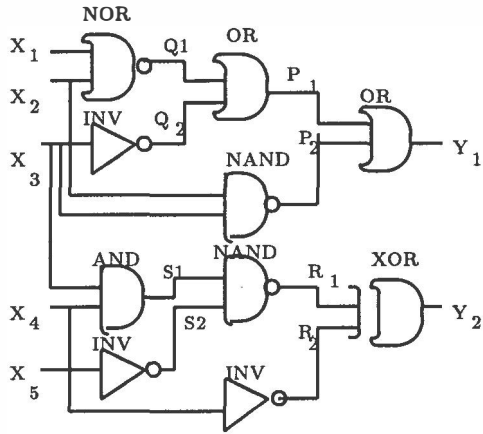

Figure 3: A circuit example.

- Assume that there is a uniform distribution on the following cases:

  i meta process    $x_2^i$, i=1 to $d_{max} + 1$.

  Hence, the expected inspection cost and replacement cost will equal

  $1/(d_{max}+1) * (x_2^1 + x_2^2 + \ldots + x_2^{d_{max}+1})$.

3. **Compute the expected time complexity and external repair cost of the bridge fault component lookahead ($Y_1$ and $Y_2$).**

# 6    An Example

We are given a circuit in figure 3 and the corresponding functional subsystem hierarchy and chip information in figure 4. We are also given a problem instance in which the input for $X_1$, $X_2$, $X_3$, $X_4$, $X_5$ is { 0, 1, 1, 1, 1}, and the observed output for $Y_1$-subsystem and $Y_2$-subsystem is {1, 1} (correct output is { 0, 1}). Therefore, we know that the highest level faulted subsystem is $Y_1$-subsystem.

Next, we briefly demonstrate the application of our method to this diagnosis problem instance.

1. **meta-level component :**

   - Compute the expected time complexities and external repair costs of the functional component lookahead and bridge fault component lookahead.
   - With the above results, evaluate the available meta-level decision model.
   - Assume the recommended action from the above evaluation is to execute the functional component.

2. **functional component:**

Figure 5 shows the result after substituting CS and NS in the conceptual decision model with the highest level faulted subsystem hierarchy. (Hereafter unbounded nodes represent chance nodes.)

- Define decision alternatives :
  testpoints : $P_1$, $P_2$.
  treatments for OR component : nothing, replace.
  treatments for $P_1$-subsystem, $P_2$-subsystem : nothing, replace, repair.
- Compute decision costs for repair treatments.
- Evaluate the decision model in figure 5. Assume the result of this evaluation is :
  Test = probe $P_1$
  - If R = $P_1$-subsystem is not ok, then Treatment = repair $P_1$-subsystem.
  - If R = $P_1$-subsystem is ok, then Treatment = repair $P_2$-subsystem.

  Figure 6 shows the result after executing the recommended actions.
- Execute next functional meta process, and assume the functional component declares failure after executing the recommended action - replacement of the OR component.

3. **meta-level component:**

   Figure 7 shows the functional component lookahead.

   Assume the recommended action after evaluating this new meta-level decision model is to execute the bridge fault component.

4. Assume the bridge fault component declares failure.

5. Repeat the meta-level component until our method locates and repairs the fault.

# 7    Related research

Other recent related work in diagnosis, uniformly treating the tasks of observation and repair, includes [16, 6, 12]. [16, 6] are both model-based diagnosis and repair systems. They begin with initial candidate sets of multiple diagnoses, generate possible next actions, choose the most utile action, update the device state and candidate sets, and loop till the device is fixed. Their work relies on a diagnosis reasoner which produces candidate sets of multiple diagnoses. They use a cost function to guide the choice of the next maximum payoff action. Our work is also a model-based diagnosis and repair system. Unlike these systems, ours is restricted to single fault, but is more uniformly decision-theoretic, needing no external diagnostic reasoner, generates the possible actions, and chooses the most utile action. Furthermore, our method allows for multiple classes (causal pathways) of errors.

[12] uses temporal influence diagrams to model the diagnosis and treatment of acute abdominal pain, and

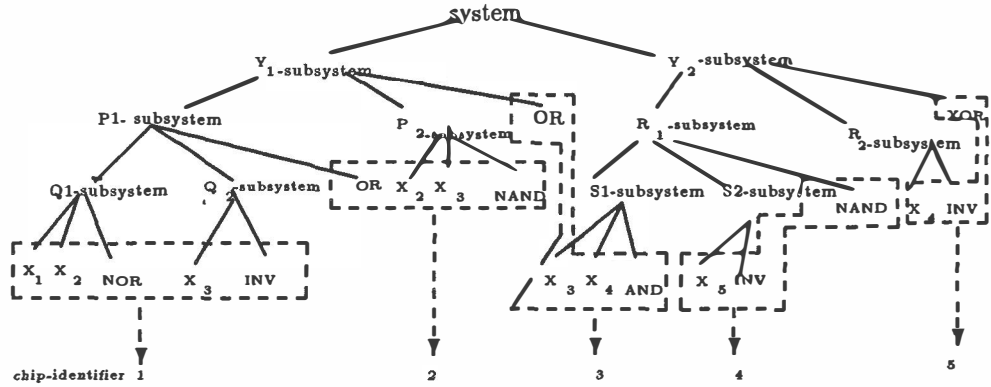

Figure 4: Functional subsystem hierarchy and chip information.

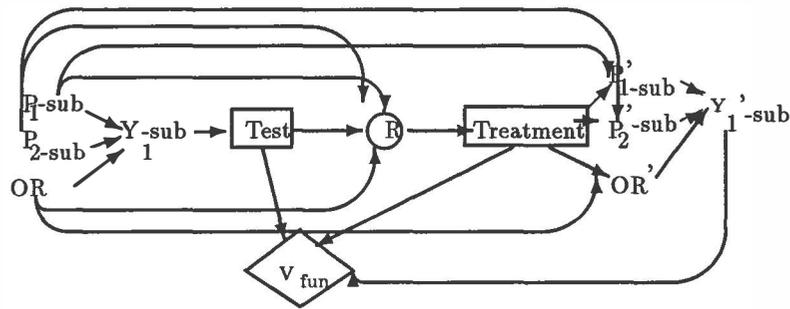

Figure 5: After substitution.

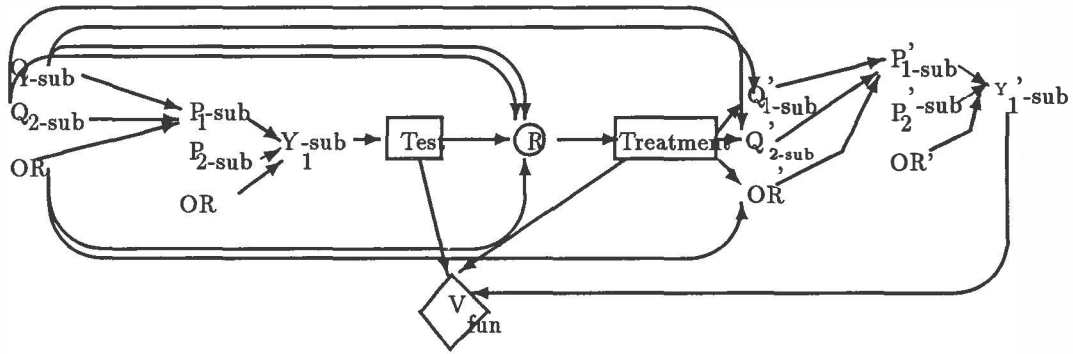

Figure 6: After executing recommended actions.

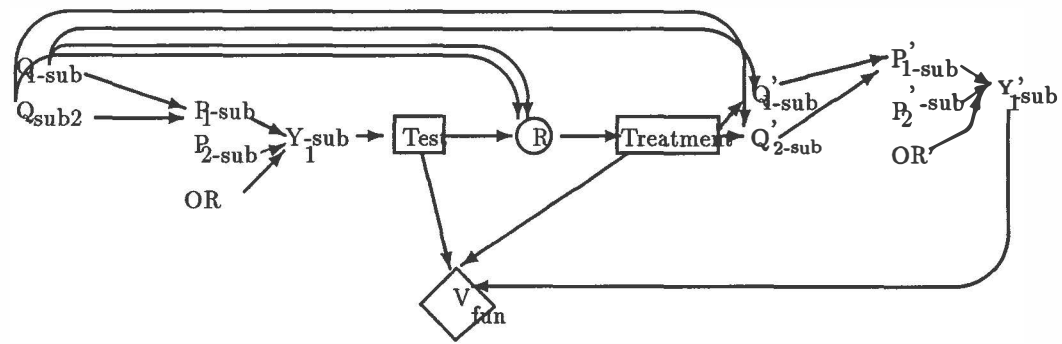

Figure 7: Functional component lookahead.

uses a dynamic influence diagram construction and updating system to automatically generate influence diagrams. The construction of influence diagrams uses a flat rule-based structure, similar to [2]. In contrast, our construction of influence diagrams uses top-down hierarchical structure and should scale more effectively.

## 8    Conclusion

This paper has described a proposed KBDMC method for resource-limited, hierarchical, complete diagnosis problems. We believe the following concepts are applicable in some other tasks of the same problem characteristic as well: (1)top-down hierarchical, incremental decision model construction, interleaved with evaluation; (2) uniform decision-theoretic treatment of meta-level and base-level tasks. Furthermore, this method only retrieves knowledge relevant to the current problem instance. That is, its knowledge use is context-sensitive.

This paper incorporates the decision-theoretic principles into hierarchical diagnostic models, treating both probe and repair as actions. With the decision-theoretic principles, the method uniformly treats the selection of causal pathways and the selection of actions in a causal pathway. Rigorous evaluation of the efficiency advantages of hierarchical refinement awaits completion of the implementation, which is in progress.

In conclusion, while there is definitely much more to be accomplished in this project, we believe we have established the essential methodology. We also show its potentials in KBDMC and diagnosis. Future work for this project includes (1) implementation of our KB-DMC system; (2) experimental evaluation of relative efficiency on diagnosis; (3) formal analysis of the task of decision model construction; (4) extension of the diagnosis problem scope.

## Acknowledgments

I would like to thank my advisor, professor Bruce D'ambrosio, for giving me valuable discussions during my research and for reviewing my earlier drafts of this manuscript.